\documentclass{article}
\usepackage{iclr2027_conference,times}

\usepackage{amsmath,amsfonts,bm}

\def\eqref#1{equation~\ref{#1}}

\def\1{\bm{1}}

\DeclareMathAlphabet{\mathsfit}{\encodingdefault}{\sfdefault}{m}{sl}
\SetMathAlphabet{\mathsfit}{bold}{\encodingdefault}{\sfdefault}{bx}{n}

\usepackage{amsmath,amssymb,amsfonts}
\usepackage{booktabs,multirow,array,tabularx,makecell}
\usepackage{graphicx}
\usepackage{microtype}
\usepackage{xspace}
\usepackage{enumitem}
\usepackage{algorithm}
\usepackage{algpseudocode}

\usepackage{xcolor}
\usepackage{url}
\usepackage{hyperref}
\definecolor{darkcerulean}{rgb}{0.03,0.27,0.49}
\definecolor{burgundy}{rgb}{0.45,0.06,0.10}
\definecolor{majorelleblue}{rgb}{0.38,0.31,0.86}
\hypersetup{
  colorlinks=true,
  citecolor=darkcerulean,
  linkcolor=burgundy,
  urlcolor=majorelleblue
}

\newcommand{\LinkTable}[1]{\hyperref[#1]{Table~\ref*{#1}}}
\newcommand{\LinkFigure}[1]{\hyperref[#1]{Figure~\ref*{#1}}}
\newcommand{\LinkSection}[1]{\hyperref[#1]{Section~\ref*{#1}}}
\newcommand{\LinkAppendix}[1]{\hyperref[#1]{Appendix~\ref*{#1}}}
\newcommand{\LinkAlgorithm}[1]{\hyperref[#1]{Algorithm~\ref*{#1}}}
\newcommand{\LinkEquation}[1]{\hyperref[#1]{Eq.~(\ref*{#1})}}
\usepackage{tikz}
\usetikzlibrary{positioning,arrows.meta,fit,calc}

\newcommand{\method}{\textsc{CounterMem}\xspace}

\title{CounterMem: World-Model-Verified Counterfactual Memory for Language Agents}

\author{
\makebox[0.28\textwidth][c]{%
\textbf{Hongji Pu}$^{1}$
}
\And
\makebox[0.28\textwidth][c]{%
\textbf{Ruixiang Tang}$^{2}$
}
\And
\makebox[0.28\textwidth][c]{%
\textbf{Yongfeng Zhang}$^{2}$
}
\AND
\makebox[0.95\textwidth][c]{%
\shortstack[c]{%
$^{1}$University of Illinois Urbana--Champaign\\
$^{2}$Department of Computer Science, Rutgers University
}}
}

\iclrfinalcopy
\iclrfinalcopy

\iclrfinalcopy
\begin{document}
\maketitle
\raggedbottom

\begin{abstract}
Existing agent memory frameworks mainly create memory through an agent’s interaction with the factual world, e.g., remembering feedback from actions taken to improve performance on future tasks. However, these frameworks seldom ask the “what if” question during memory construction: what if a different action had been taken, would the feedback have changed, and how could this feedback become useful memory? Obtaining such feedback directly in an active environment can be expensive and can alter the state needed for comparison. In this work, we introduce \method, a reinforcement-learning framework for constructing and using verified counterfactual memory across tasks. After a failed action, \method evaluates local alternatives from a copy or reset of the original state using executable world models, such as tests, proof checkers, and solvers. It stores improvements with the original and corrected actions, checked outcomes, and conditions for reuse. A learned memory-use policy selects a retrieved record or skips memory to balance task success and interaction cost, while the base LLM remains fixed. Both memory and policy are frozen during held-out evaluation. We evaluate \method on 12 benchmark settings across six domains. With gpt-oss-120b, \method improves both ReAct and Reflexion on all
12 benchmarks across six domains, averaging a gain of 12.6 percentage
points over their unaugmented versions. In the four-domain comparison
across two backbones, task-run tokens decrease by 7.7--42.0\%, excluding
offline selector-training costs. Further analyses show that removing
verification or persistent storage weakens the gains, while applying
verified corrections to unsuitable decisions can reverse them. Code will be released upon acceptance.

\end{abstract}

\section{Introduction}
\label{sec:introduction}

Agent memory turns interaction into experience that can guide future tasks.
A coding agent remembers test feedback, a theorem-proving agent retains
proof attempts, and a database agent learns from query results.
For example, Reflexion stores verbal reflections on these outcomes
\citep{shinn2023reflexion}; ExpeL extracts reusable insights from trajectories
\citep{zhao2023expel}. Existing memory mechanisms mainly describe factual trajectories:
the actions an agent took and the feedback it received. Retaining this
experience can reduce the need to rediscover a useful solution.

The same trajectory leaves another source of experience unexplored:
\emph{what if the agent had taken a different action at that decision?}
Observing that action $A$ failed does not establish whether action $B$
would have improved the outcome. A reflection may propose $B$, but without
checking it, the agent may miss the opportunity of storing a plausible suggestion as useful
experience. The solution is to construct memory from evaluated
alternatives, so that later tasks can benefit from corrections absent
from the original trajectory.

However, obtaining this counterfactual feedback is not always straightforward.
Trying alternatives in an active environment can consume additional
interactions or alter the state needed for comparison. Even when actions
can be replayed, evaluating every alternative is costly. Many computational
tasks offer a more controlled setting: repository tests check patches
\citep{jimenez2024swebench}, proof assistants check proofs
\citep{zheng2022minif2f}, and SQL benchmarks check query results
\citep{li2023bird}. When the decision state can be copied or reset, these
tools can evaluate a bounded set of alternatives without committing them
to the original trajectory. We use such executable evaluators as world models to enable counterfactual evaluation.

\begin{figure}[t]
\centering
\includegraphics[width=0.94\linewidth]{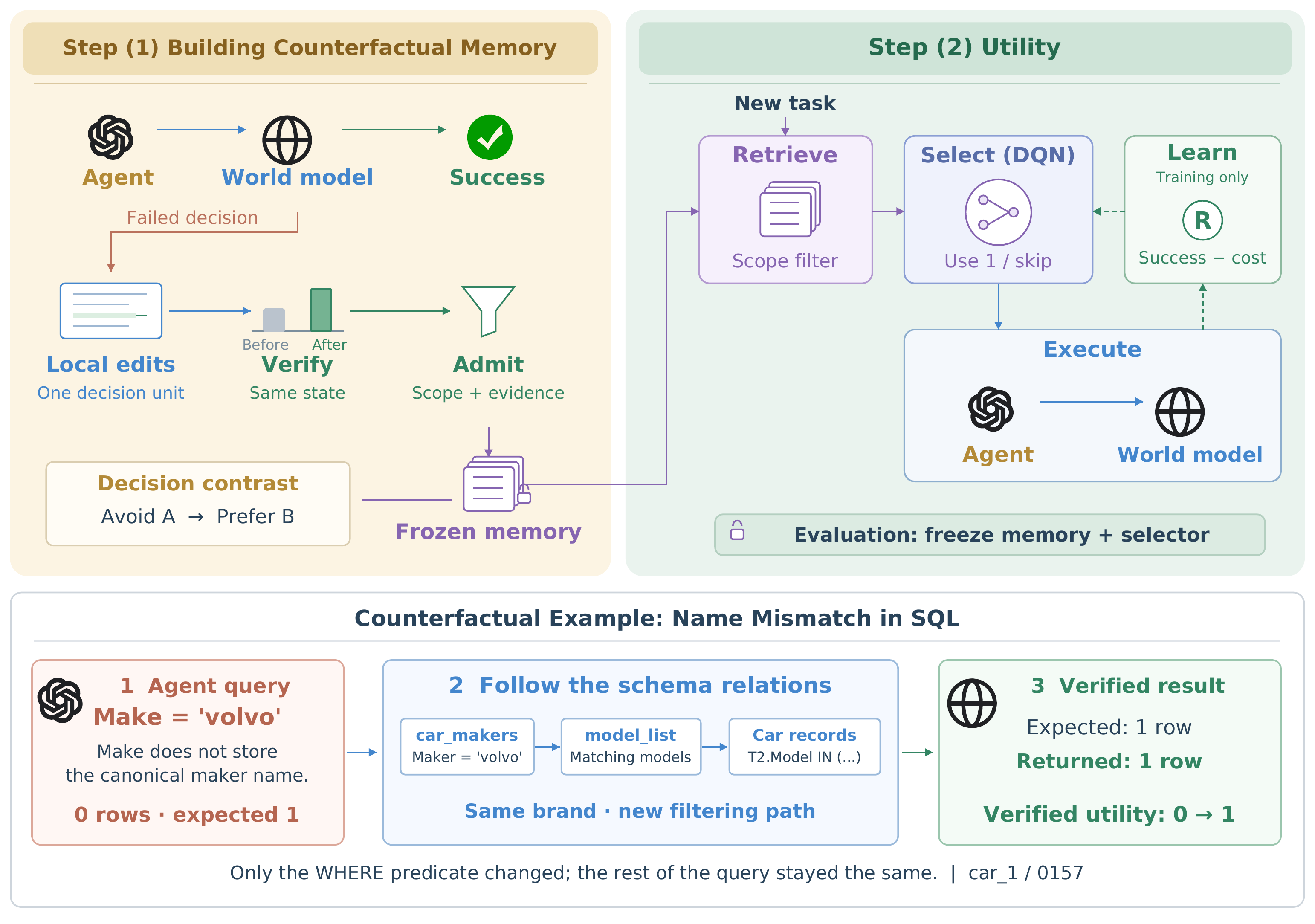}
\caption{\textbf{Constructing and using counterfactual memory.}
Local alternatives are checked from a copy or reset of the original state.
Verified improvements become records containing the action contrast,
evidence, and applicability conditions. On later tasks, retrieval supplies
compatible records, and a trained DQN selector chooses one or skips memory.
Memory and selector parameters are frozen during held-out evaluation.}
\label{fig:countermem}
\end{figure}

Counterfactual feedback becomes more useful across tasks when the agent
preserves what changed and when the correction applies. A correction
verified on one task may not apply to a similar task with different
conditions. This motivates two decisions: \emph{which evaluated
alternatives should become memory, and when should that memory guide
a later action?}

Motivated by the above problems,  we introduce \method, a reinforcement-learning framework for constructing
and using verified counterfactual memory across tasks
(\LinkFigure{fig:countermem}). After a failed decision, it proposes local
edits, evaluates them from the same state, and stores improvements with
their action contrasts, checked outcomes, and applicability conditions.
A memory-use policy learns to select a retrieved record or skip memory,
with rewards balancing task success and interaction cost. This assigns
credit to useful reuse rather than assuming that every verified record
will help. The base LLM remains fixed. Construction and policy training
exclude evaluation tasks, and both memory and policy are frozen for
held-out evaluation.

With gpt-oss-120b, \method improves both agents on all 12 benchmarks
across six domains, averaging +12.6 points. Across two backbones, it
reduces task-run tokens by 7.7--42.0\%. Further studies show that effective
reuse requires retaining useful records and matching verified corrections
to appropriate decisions.
Our contributions are:

\begingroup
\setlength{\parskip}{0pt}
\begin{itemize}[leftmargin=*,itemsep=0pt,parsep=0pt,topsep=2pt,partopsep=0pt]
\item \textbf{Constructing memory beyond factual trajectories.}
We turn checked local alternatives into persistent action contrasts with
supporting evidence and conditions for reuse (\LinkSection{sec:method}).
\item \textbf{Learning when counterfactual memory is useful.}
We formulate record selection and skipping as an RL problem, linking
source-task evidence to later success and interaction cost
(\LinkSection{sec:memory_rl}).
\item \textbf{Testing memory quality as well as task performance.}
Experiments measure gains, accumulation, and transfer; interventions
examine the roles of verification, state matching, and applicability
conditions (\LinkSection{sec:results}; \LinkSection{sec:analysis}).
\end{itemize}
\endgroup

\section{Related Work}
\label{sec:related}

\textbf{Agent memory, reflection, and counterfactual experience.}
Agent memory retains reflections, experiences, workflows, and reusable
strategies across tasks, as in Reflexion \citep{shinn2023reflexion},
ExpeL \citep{zhao2023expel}, Generative Agents
\citep{park2023generative}, Voyager \citep{wang2023voyager}, MemGPT
\citep{packer2023memgpt}, Agent Workflow Memory \citep{wang2025awm},
A-Mem \citep{xu2025amem}, and ReasoningBank
\citep{ouyang2026reasoningbank}. Counterfactual reasoning and search
instead evaluate alternative outcomes or actions
\citep{vashishtha2026counterfactuals,antoniades2025swesearch}.
\method connects these directions by evaluating local action alternatives
at the same decision state and retaining their checked outcome contrasts
and applicability conditions for reuse across future tasks.

\textbf{World models and executable feedback.}
World models support decision making by predicting or simulating action
consequences for planning \citep{hafner2019learning,hafner2020dreamer}.
Recent agent systems extend this idea through language-based environment
models \citep{chae2025web} and executable simulators
\citep{lehrach2026code}, while executable tests provide direct feedback
for software agents \citep{jimenez2024swebench}. \method uses
task-specific tests, checkers, and solvers as executable world models to
compare local alternatives at the same decision state and preserve
verified contrasts as cross-task memory. Unlike current-task planning,
these comparisons are retained with conditions for later reuse, while
their evidence remains bounded by the tested state and available checks.
\section{Methodology}
\label{sec:method}

\subsection{What the Agent Remembers}

\method saves tested corrections from earlier tasks to help an agent
solve later ones (\LinkFigure{fig:countermem}). The \emph{base LLM} writes
actions, such as SQL queries or code edits, and an \emph{evaluator} checks
their results. When an edit improves a failed action, the \emph{memory
store} can save what changed, the check results, and the conditions for
using that correction. On a later task, a \emph{selector} chooses one
stored record or skips memory. The LLM uses the selected record to revise
its draft action before execution.

We first build memory on \emph{source tasks}, which supply the corrections.
We then fix the store and train the selector to choose when to use those
records. During testing, both the store and selector remain fixed.
The base LLM's parameters remain unchanged throughout.

For example, a query for people aged \emph{at least} 18 may incorrectly use
\texttt{age > 18}. \method tests \texttt{age >= 18} from the same database
state. If the result checker confirms improvement, the method can save
the original and corrected predicates, their results, and the requirement
to include the boundary value. A later ``at least'' query can use this
record to correct the same mistake. A query for ages \emph{strictly above}
18 should keep the strict comparison. Modules I--II build these records;
Module III retrieves them and decides whether they should guide a new action.

\subsection{Module I: Propose Alternatives to a Failed Action}
\label{sec:generation}

During memory construction, the agent takes action $a_t$ in state $s_t$,
which includes the current task, environment, partial solution, and
feedback so far. The evaluator $H$ returns the outcome
$o_t=H(s_t,a_t)$. We use the term \emph{world model} for this evaluator
because it checks what happens when an action is tried. Depending on the
task, $H$ is a test suite, proof checker, SQL result checker, or solver.
Its feedback comes from executing these checks.

When an action fails and checking budget remains, a generator proposes
at most $K=4$ local alternatives:
\begin{equation}
\mathcal A_{\mathrm{cf}}=G(s_t,a_t,o_t;K).
\label{eq:generator}
\end{equation}
The generator $G$ uses LLM edits or rules for the observed failure to
change one part of the action, such as a SQL condition, proof step, or
code expression. Each resulting action is a candidate for evaluation.
These alternatives ask a \emph{counterfactual} question: what would have
happened if the agent had taken this edited action instead?

\subsection{Module II: Check Alternatives and Store Corrections}
\label{sec:verification}

Each alternative $a'$ is evaluated from a copy or reset of $s_t$.
This lets us compare the edited and original actions from the same
starting state:
\begin{equation}
\Delta(a')=U\!\left(H(s_t,a')\right)-U(o_t).
\label{eq:cf_gain}
\end{equation}
$U$ converts evaluator feedback to a score in $[0,1]$, such as correctness
or the fraction of tests passed. Thus, $\Delta$ measures the improvement
due to the edit. A candidate qualifies for storage only if its check
completes and $\Delta(a')>\epsilon_{\mathrm{adm}}\geq0$, where
$\epsilon_{\mathrm{adm}}$ is the minimum required gain. An improved
partial score may qualify even if the whole task remains unsolved.

An accepted correction becomes a record $m=(q,a^{-},a^{+},e,c,u)$.
The record describes the situation $q$, the original action $a^{-}$,
the improved action $a^{+}$, and their check results $e$.
The field $c$ states the task requirements under which the correction
applies. In the SQL example, $q$ describes a missed boundary value,
while $c$ requires an inclusive request such as ``at least'' and
compatible column types. Retrieval uses $q$ to find related records
and checks $c$ against the new task before offering them for use.
The evidence $e$ compares query results; successful execution alone
does not establish improvement.

The remaining field $u$ summarizes what happened when this record was
reused on subsequent source tasks during construction. It starts at zero
and is updated when reuse outcomes are observed. These observations
precede selector training: once construction ends, the records and their
reuse statistics are fixed. We also retain each record's source-task
identifier so that retrieval can exclude records from the current task.

The store holds at most $B=200$ records in the specified configuration.
Admission rules check proposed records for duplicate or conflicting
advice. When space is needed, retention rules decide which records to
remove (\LinkAppendix{app:math}). Saving a correction makes it available
for later tasks; it does not automatically replace the action in its
source task.

\subsection{Module III: Retrieve a Record and Decide Whether to Use It}
\label{sec:memory_rl}

On a later task, the LLM first drafts an action $\hat a_t$.
Retrieval excludes records from the same task and filters out records
whose conditions $c$ do not fit the current task and environment.
It ranks the remaining records by the relevance of their situation $q$
and their stored reuse statistic $u$, returning at most $k=3$ candidates.

The selector decides whether one of these candidates would help revise
the current draft. A related record may be unnecessary: for example,
the draft may already include the requested boundary value.
If the selector chooses a record, the LLM adapts its correction to the
draft. If the selector skips memory, the draft proceeds unchanged.
With no eligible record, it must skip. The evaluator checks the resulting
action in either case.

We train the selector as a deep Q-network (DQN) \citep{mnih2015human}.
At each training decision, it receives the current state, draft,
candidate records, and remaining budgets. Its choices are to supply one
candidate or skip memory. Training rewards task completion and penalizes
evaluator calls, failed attempts, and token use, teaching the selector
to score these choices by their expected reward.

The three quantities serve different purposes: $\Delta$ measures how much
the original edit helped on its source task; $u$ summarizes observed reuse
during construction; and the selector predicts the reward from using a
record at the current decision. \LinkAppendix{app:selector} gives the
reward and training rules. During testing, the fixed selector chooses
the valid option with the highest predicted reward.
\section{Experimental Setup}
\label{sec:setup}

We compare memory mechanisms, then vary source experience, record retention,
and the tasks on which memory is used. We use success and cost to measure whether earlier corrections help solve
new tasks.

\subsection{Tasks, Agents, and Memory Baselines}

\textbf{Tasks and checks.}
The suite contains 12 benchmark settings in six domains: mathematics,
coding, formal verification, text-to-SQL, SMT, and SAT. A task asks the
agent to produce a proof, repair, verified program, query, or solver
solution. Lean checks proofs \citep{demoura2021lean4}; repository tests
check repairs \citep{jimenez2024swebench}; Dafny and Verus check programs
\citep{leino2010dafny,lattuada2023verus}; SQL results are compared with
reference results \citep{li2023bird}; and SMT/SAT solvers check solver
inputs and outcomes \citep{barrett2016smtlib,hoos2000satlib}.
\LinkAppendix{app:benchmarks} maps each benchmark to its checker.

\textbf{Main comparison.}
\LinkTable{tab:multiagent_multimodel} crosses ReAct \citep{yao2023react}
and Reflexion \citep{shinn2023reflexion} as base agents with gpt-oss-120b
\citep{openai2025gptoss} and Gemini 3.1 Flash-Lite
\citep{google2026gemini31flashlite} as base LLMs. Each group compares the following memory mechanisms: no augmentation,
ExpeL, Generative Agents, Voyager, MemGPT, and CF Memory; Best-of-$N$
provides a separate compute control. Within each group, comparing a row with ``None'' measures its gain over the base agent; comparing augmented rows tests the different added memory
mechanisms. Table~1 reports four-domain results for Math, Coding,
Text-to-SQL, and SAT; the full 12-benchmark evaluation across six domains,
including formal verification and SMT, is reported in \LinkAppendix{app:taskvalues}.

The base agent is the procedure for using the LLM to solve tasks.
ReAct asks the LLM to propose actions and then respond to tool feedback.
Reflexion also asks it to write reflections on earlier attempts and use
them in later attempts. ``None'' adds no memory mechanism and keeps these
reflections. CF Memory adds tested corrections, even when reflections
are already available.

\textbf{What the added mechanisms provide.}
ExpeL supplies lessons from past attempts \citep{zhao2023expel}; Generative
Agents supplies stored experiences and reflections \citep{park2023generative};
Voyager supplies reusable procedures \citep{wang2023voyager}; MemGPT manages
which information stays in context or external storage \citep{packer2023memgpt}.
We adapt these memory components to a common task interface. CF Memory
supplies checked corrections through the full \method process: build
records, retrieve candidates, and choose one or skip. Best-of-$N$ generates
extra candidates for the current task
without keeping memory, asking whether extra trials can match reuse.
ReasoningBank extracts strategies from past attempts
\citep{ouyang2026reasoningbank}; it is the memory baseline in the
accumulation, transfer, and fixed-budget studies.

\begin{table*}[t]
\centering
\caption{\textbf{Four-domain comparison of memory mechanisms across
agents and backbones.}
Scores (0--100, higher is better) average benchmarks within each domain;
tokens report total task-run consumption in millions. Bold marks the best
value within each agent--backbone group.}
\label{tab:multiagent_multimodel}

\scriptsize
\setlength{\tabcolsep}{3.5pt}
\renewcommand{\arraystretch}{1.08}

\begin{tabular*}{\textwidth}{
@{\extracolsep{\fill}}
llccccc
@{}
}
\toprule
\textbf{Base Agent}
&
\textbf{Augmentation}
&
\shortstack{\textbf{Math}\\$\uparrow$}
&
\shortstack{\textbf{Coding}\\$\uparrow$}
&
\shortstack{\textbf{Text-to-SQL}\\$\uparrow$}
&
\shortstack{\textbf{SAT}\\$\uparrow$}
&
\shortstack{\textbf{Tokens}\textbf{(M)}\\ $\downarrow$}
\\
\midrule

\multicolumn{7}{l}{
\textit{\textcolor{gray}{Backbone:
\href{https://developers.openai.com/api/docs/models/gpt-oss-120b}
{gpt-oss-120b}}}
}
\\
\midrule

\multirow{7}{*}{ReAct}
& None
& 29.8 & 48.7 & 26.8 & 50.7 & 6.0 \\

& Best-of-$N$ (compute control)
& 34.5 & 38.1 & 36.2 & 41.1 & 9.4 \\

& ExpeL \citep{zhao2023expel}
& 27.4 & 56.5 & 17.2 & 49.0 & 5.4 \\

& Generative Agents \citep{park2023generative}
& 22.6 & 37.1 & 17.6 & 41.6 & 6.2 \\

& Voyager \citep{wang2023voyager}
& 21.9 & 60.6 & 26.3 & 51.4 & 5.2 \\

& MemGPT \citep{packer2023memgpt}
& 27.3 & 39.5 & \textbf{48.9} & 47.0 & 5.8 \\

& \textbf{CF Memory}
& \textbf{57.5} & \textbf{66.6} & 47.0 & \textbf{58.9} & \textbf{4.4} \\

\midrule

\multirow{7}{*}{Reflexion}
& None
& 40.8 & 34.6 & 54.6 & 61.2 & 5.2 \\

& Best-of-$N$ (compute control)
& 31.8 & 21.7 & 62.5 & 53.6 & 10.0 \\

& ExpeL
& 38.0 & 32.8 & \textbf{68.1} & 60.0 & 5.0 \\

& Generative Agents
& 36.7 & 38.8 & 60.4 & 48.3 & 5.6 \\

& Voyager
& 37.7 & \textbf{42.6} & 55.4 & 58.3 & \textbf{4.5} \\

& MemGPT
& 48.0 & 39.1 & 47.5 & 58.0 & 5.7 \\

& \textbf{CF Memory}
& \textbf{53.4} & 40.7 & 66.1 & \textbf{63.7} & 4.8 \\

\midrule
\midrule

\multicolumn{7}{l}{
\textit{\textcolor{gray}{Backbone:
\href{https://ai.google.dev/gemini-api/docs/models/gemini-3.1-flash-lite}
{Gemini 3.1 Flash-Lite}}}
}
\\
\midrule

\multirow{7}{*}{ReAct}
& None
& 38.5 & 30.9 & 57.6 & 45.0 & 5.0 \\

& Best-of-$N$ (compute control)
& 30.7 & 29.0 & 67.3 & 34.9 & 9.6 \\

& ExpeL
& \textbf{57.4} & 37.3 & 68.6 & 37.2 & 4.8 \\

& Generative Agents
& 42.4 & 31.3 & 57.2 & 43.9 & 5.7 \\

& Voyager
& 39.9 & 34.4 & 48.2 & 55.1 & 4.3 \\

& MemGPT
& 31.4 & 38.2 & 68.5 & 45.2 & 5.5 \\

& \textbf{CF Memory}
& 54.5 & \textbf{40.0} & \textbf{71.7} & \textbf{57.9} & \textbf{2.9} \\

\midrule

\multirow{7}{*}{Reflexion}
& None
& 55.5 & 47.7 & 59.1 & 43.7 & 4.3 \\

& Best-of-$N$ (compute control)
& 58.7 & 43.9 & 65.6 & 54.7 & 8.2 \\

& ExpeL
& 50.3 & 43.2 & 70.2 & 36.8 & 4.0 \\

& Generative Agents
& 44.6 & 36.6 & 66.6 & \textbf{66.1} & 4.9 \\

& Voyager
& 54.9 & 37.1 & 49.0 & 32.1 & 3.9 \\

& MemGPT
& 43.8 & 38.2 & 52.0 & 49.5 & 4.8 \\

& \textbf{CF Memory}
& \textbf{74.0} & \textbf{73.5} & \textbf{77.8} & 62.3 & \textbf{3.4} \\

\bottomrule
\end{tabular*}
\end{table*}

\subsection{Build Memory, Train Selection, Then Test}
\label{sec:heldout}

\textbf{Construction tasks produce the stored records.}
On source tasks, the agent tries actions; after a failure, \method tests
local edits and saves accepted corrections, as in Modules I--II.
The environment and temporary working state reset between tasks; saved
records carry corrections forward, and their reuse statistics can be updated
during this stage. Each benchmark, LLM, agent, configuration, and seed
has its own store.

\textbf{Training tasks teach the selector to use that store.}
The store, including reuse statistics, is now fixed. The DQN tries record
choices or skipping, observes task success and cost, and updates only its
selection parameters. Training uses 2,000 decisions.
\LinkAppendix{app:selector} gives the training settings and
task-set separation.

\textbf{Held-out tasks measure reuse of earlier experience.}
Held-out tasks appear in neither construction nor selector training.
The agent can use feedback to revise its answer within a test task, but
records, reuse statistics, and selector parameters stay fixed across test
tasks. Success therefore measures use of experience collected before
testing. \LinkAppendix{app:heldout} gives the SQL split example.

\subsection{What the Additional Experiments Change}

\textbf{Accumulation and retention.}
We compare \method and ReasoningBank after 0, 25, 50, 100, and 200 source
tasks, evaluating each saved store on held-out tasks. Checkpoints count
processed tasks rather than stored records: one task may yield zero or
multiple records, and records can be replaced once capacity is reached.
We separately compare recent, random, similarity-ranked, and utility-ranked
retention at equal record capacity. \LinkAppendix{app:accumulation_protocol}
gives further details.

\textbf{Transfer and ablations.}
We compare \method and ReasoningBank under four changes to the next task:
a new request in the same environment, a request in a different environment,
the same error expressed differently, and a look-alike request that makes
the old correction wrong. An environment is the execution setting, such
as a database or repository. For a look-alike example, changing \texttt{>}
to \texttt{>=} is wrong when the new request says ``strictly above.''
We measure success, whether guidance is retrieved, and harmful use.
Ablations separately remove checks or storage, or change which records
reach the selector (\LinkSection{sec:analysis}).

\subsection{Evaluation Metrics}

Scores use benchmark success measures on a 0--100 scale; differences are
percentage points (pp). Domain columns average their benchmarks.
Four-domain summaries weight domains equally; 12-benchmark summaries
weight benchmarks equally.

Total tokens count reasoning, retrieval, and augmentation in the reported
task runs, in millions (M). Evaluator calls include counterfactual checks;
failed attempts are evaluated actions that fail the success criterion.
Interaction plots divide total calls or failed attempts by the number of
solved tasks; a lower ratio need not mean fewer total calls. Accumulation panel
(b) counts only retrieval and context tokens per task. Offline selector
training is not separately costed. Best-of-$N$ has a different token total;
a separate 6M-token comparison appears in \LinkAppendix{app:fixedbudget}.
Transfer-rate definitions and reporting limits are in
\LinkAppendix{app:transfer_protocol}.

\section{Results}
\label{sec:results}

\subsection{Does Earlier Experience Help Solve New Tasks?}

Within each agent--LLM group in \LinkTable{tab:multiagent_multimodel},
we first compare CF Memory with ``None.'' For example, averaging the four
domain columns for gpt-oss-120b with ReAct gives 39.0 without augmentation
and 57.5 with CF Memory, a gain of 18.5pp. Gains across four groups
range from 8.2 to 20.4pp. Reflexion gains 8.2pp with gpt-oss-120b and
20.4pp with Gemini 3.1 Flash-Lite, so the added corrections help even when
reflections are available. Comparing all rows within each group, \method
has the highest four-domain mean in all four groups and leads in 11 of
16 domain comparisons.

Task-run tokens fall by 7.7--42.0\% relative to ``None''.
Best-of-$N$ uses 8.2--10.0M tokens, compared with 2.9--4.8M for \method,
yet scores lower on average in every combination. The reported Best-of-$N$ control therefore does not match \method's mean score, despite using more task-run tokens.

\subsection{What Changes as the Agent Encounters More Source Tasks?}
\label{sec:accumulation}

\begin{figure}[t]
\centering
\includegraphics[width=\linewidth]{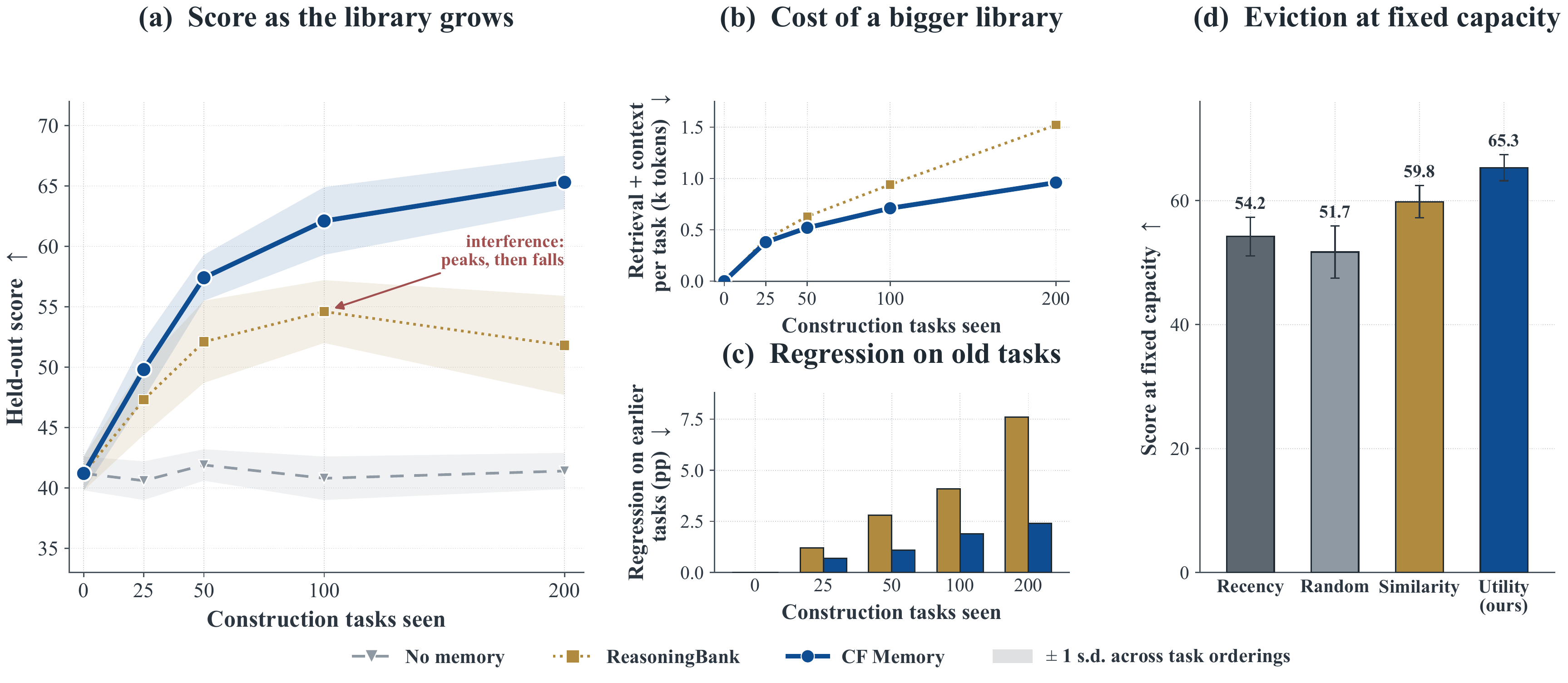}
\caption{\textbf{Learning from more source tasks and choosing what to keep.}
(a) Test score at successive memory-construction checkpoints.
(b) Retrieval and context tokens per task; other solving costs are excluded.
(c) Reported score loss on earlier tasks.
(d) Test score under different record-retention rules at equal capacity.
The horizontal axis in (a--c) counts source tasks, not stored records;
shading shows one standard deviation across task orderings.}
\label{fig:accumulation}
\end{figure}

Moving right in \LinkFigure{fig:accumulation}(a) means testing memory
collected after more source tasks. \method's held-out score rises, while
ReasoningBank's first rises and then falls. The latter decline shows that
processing more source tasks may not necessarily make the stored guidance more useful. Retrieval and
context costs rise for both (b). \method also loses less performance on
earlier tasks at each nonzero checkpoint (c), though neither fully preserves it. \LinkAppendix{app:accumulation_protocol}
gives the checkpoint protocol and reporting details.

Panel (d) asks a different question: with the same record limit, which
records should be kept? The utility rule favors corrections with larger
checked improvements and more helpful reuse during construction. It scores
65.3, versus 59.8 for similarity, 54.2 for recency, and 51.7 for random
retention. Its 5.5-point advantage over similarity supports utility-aware retention
when storage is limited; the advantage does not come from a larger
record capacity.

\subsection{When Does a Stored Correction Apply to a New Task?}
\label{sec:transfer}

\begin{figure}[t]
\centering
\includegraphics[width=\linewidth]{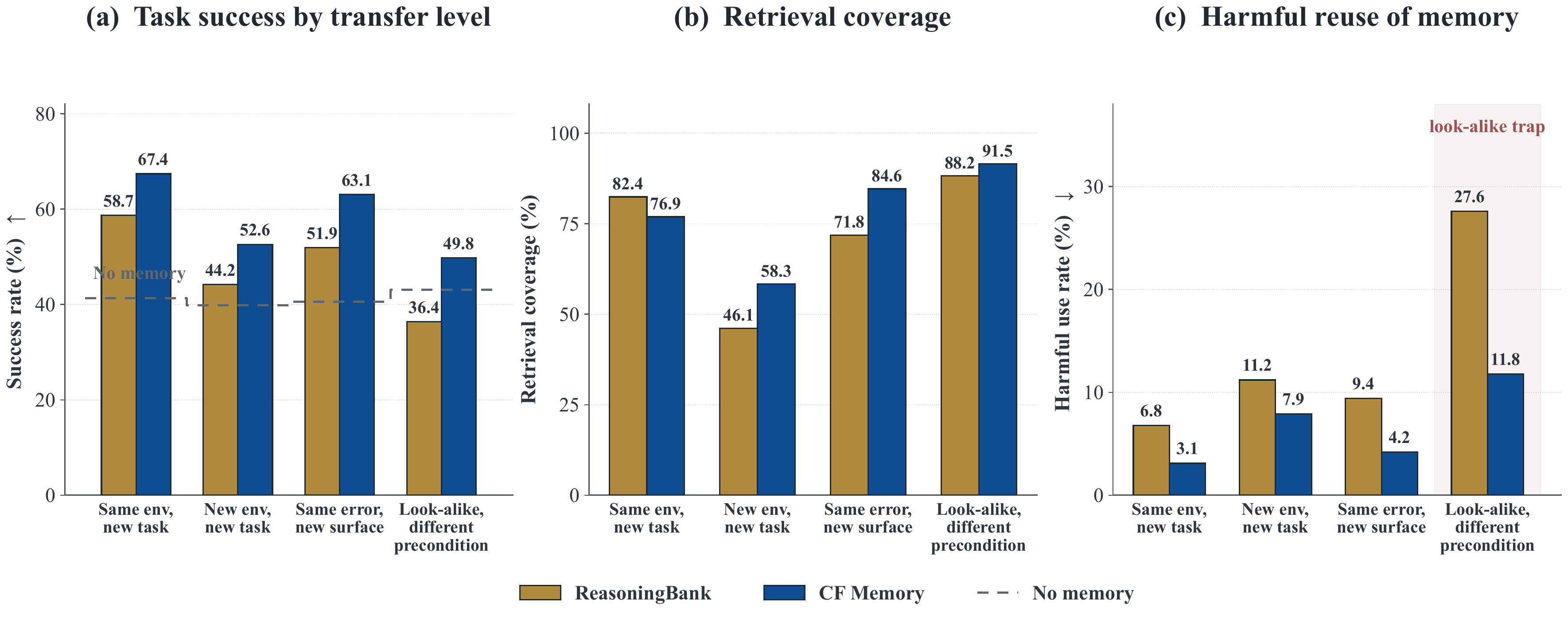}
\caption{\textbf{Reusing memory under four changes to the task.}
(a) Task success, with a no-memory reference.
(b) Frequency of obtaining retrieved guidance.
(c) Reported harmful-use percentages.
``New surface'' changes how the same error appears; ``look-alike'' preserves
similarity while changing the condition needed for the correction.
Definitions and reporting details are in
\LinkAppendix{app:transfer_protocol}.}
\label{fig:transfer}
\label{fig:transfer_full}
\end{figure}

\LinkFigure{fig:transfer} separates finding guidance (coverage), solving
the task (success), and worsening an outcome through memory use (harm).
Within each of the four conditions, \method has higher success than
ReasoningBank. Across conditions, its own score falls from 67.4 in the
original environment to 52.6 in a new one, which indicates that earlier corrections remain
useful, but their benefit does not remove the transfer difficulty. In the
original environment, higher success accompanies lower coverage
(76.9\% versus 82.4\%), meaning that finding guidance more often is insufficient
to explain which method succeeds more often.

The look-alike tasks test whether the agent avoids a familiar correction
when the new request makes it wrong, as in the SQL example in
\LinkSection{sec:method}. Both methods have high reported coverage
(88.2\% for ReasoningBank; 91.5\% for \method), but \method achieves 49.8\%
success versus 36.4\%. Reported harmful use falls from 27.6\% to 11.8\%.
To test the role of the stored condition, a separate intervention removes
the applicability field $c$ from \method records. Harmful use rises from
11.8\% to 34.2\% (\LinkAppendix{app:rescue}). This experiment supports the use of explicit
conditions to avoid applying a correction to the wrong request. 

\subsection{Does Better Performance Require Fewer Evaluations?}

For each of 12 benchmarks with gpt-oss-120b, we compare ReAct and Reflexion
with their own CF Memory versions. Scores improve in all 24 comparisons
(mean +12.6 pp; \LinkFigure{fig:domain} in
\LinkAppendix{app:taskvalues}). Costs vary: SWE-bench Verified uses
3.6/1.2 fewer calls per solved task for ReAct/Reflexion, while
Vericoding-Verus with ReAct gains 9.6pp but uses 0.5 more calls.
Thus, the score gains are consistent across benchmarks, whereas
interaction-cost changes are benchmark dependent.

\section{Further Analysis}
\label{sec:analysis}

We first change how corrections are checked and saved, then change which
decisions receive them. All comparisons use gpt-oss-120b; paired numbers
report ReAct followed by Reflexion. Score gains use each agent's own
unaugmented version as the reference.

\subsection{What Do Verification and Persistent Storage Contribute?}

\begin{figure}[t]
\centering
\includegraphics[width=0.97\linewidth]{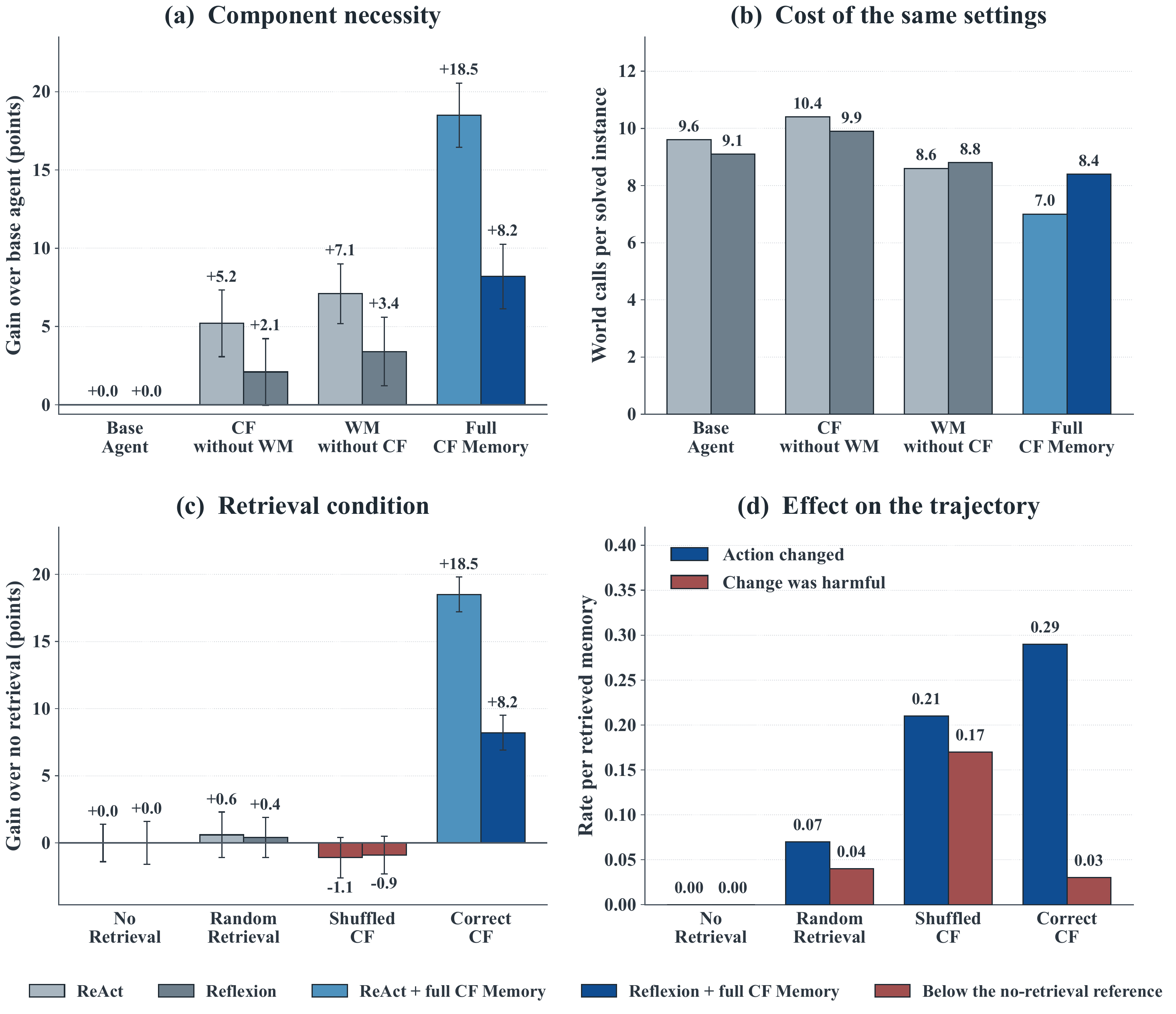}
\caption{\textbf{Removing components and changing record assignment.}
(a) Score gains over the base agent. ``CF without WM'' stores unverified
advice; ``WM without CF'' checks current-task alternatives without keeping
cross-task corrections. (b) Evaluator calls per solved task.
(c) Score gains when candidate records are absent, random, shuffled across
states, or matched to the current decision. (d) Action changes and harmful
changes per retrieved record.}
\label{fig:components}
\label{fig:intervention}
\end{figure}

``CF without WM'' removes world-model-based checks before storage, saving proposed corrections
as untested advice. ``WM without CF'' checks alternatives within the current
task but saves no correction records for later tasks. These comparisons
ask whether suggestions without checks, or checks without storage, recover
the full gain. Both variants still evaluate ordinary task actions.

Unverified advice gives +5.2/+2.1pp for ReAct/Reflexion; testing alternatives
without saving records gives +7.1/+3.4pp. The full method reaches +18.5/+8.2pp
(\LinkFigure{fig:components}(a)). Unverified advice also uses more calls
per solved task than the base agents: 10.4/9.9 versus 9.6/9.1. The full
method uses 7.0/8.4 (panel (b)). Neither untested advice nor unsaved
alternatives recover the full gain. This experiment supports our design of both checking corrections
and retaining them for reuse. 

\subsection{What Happens When a Valid Record Reaches the Wrong Decision?}

We vary where stored corrections are offered. Random retrieval ignores
state matching, shuffled retrieval reassigns records to other decision
states, and matched retrieval uses the stored situation and conditions.
Thus, a record can remain valid on its source task yet be unsuitable for
the current decision. Score changes are relative to no retrieval.

Random retrieval gives only +0.6/+0.4pp and shuffling -1.1/-0.9pp,
versus +18.5/+8.2pp with matched retrieval (panel (c)). In panel (d),
shuffling yields 0.21 action changes and 0.17 harmful changes per retrieved
record, compared with 0.29 and 0.03 for matched retrieval. Matched records
therefore change actions more often but cause less harm, supporting the
importance of matching guidance to the current decision rather than merely
limiting how often memory changes actions. See
\LinkAppendix{app:transfer_protocol}.

\section{Conclusion}
\label{sec:conclusion}

\method turns tested corrections from earlier tasks into guidance for
later ones. It checks edits to failed actions, saves what worked better
and when it applies, and trains a selector to offer one record to the LLM
or let it act without memory. 
Across 12 benchmarks in six domains, \method improves both base agents,
averaging +12.6 points with gpt-oss-120b; across two backbones, it reduces
task-run tokens by 7.7--42.0\%. Further analyses show that useful
counterfactual memory requires both verified corrections and appropriate
reuse: mismatched records can erase gains or cause harm.

\clearpage
\section{AI Use Statement(Does not count towards the page limit)}
\label{app:ai_use}

The authors used ChatGPT to revise the manuscript's language, clarify
explanations, and reduce repetition. The authors reviewed the revised
content and take responsibility for the submitted manuscript. For manuscript editing, the models were used primarily for
language polishing and concise clarification, for example by replacing
awkward phrasing with shorter equivalent expressions, adding brief
descriptive phrases such as ``as an LLM agent'' when needed for readability,
or rewriting a sentence without changing its technical meaning. The models
were not used as a substitute for experimental validation or scientific
judgment. All AI-assisted outputs were reviewed by the authors, and the
reported experimental configurations and results were checked against the
authors' run artifacts and checkpoints. The authors take responsibility for
the final content of the paper, including all claims, results, code, and
AI-assisted text or artifacts.

\section{Ethics Statement(Does not count towards the page limit)}
\label{app:ethics}

Persistent memory can carry an inaccurate or outdated correction into a
later task. Checking that correction on its source task does not establish
that it is safe to reuse under different requirements or permissions.
Records should retain their source and scope, and systems should protect
sensitive information contained in stored trajectories. The benchmark
results measure task performance; they are not a deployment safety guarantee.

\section{Reproducibility Statement(Does not count towards the page limit)}
\label{app:reproducibility}

This appendix records the configuration used by the reported runs at the
level needed to interpret the method and diagnostic figures. In particular,
it specifies the memory thresholds and retention weights, the selector
architecture and optimizer settings, the five selector seeds and checkpoint rule, the fixed-selector policy used across
accumulation snapshots, the transfer sample counts and rate denominators,
the $c$-removal intervention, and the prompt/parser interfaces. These
settings were checked against the run checkpoints used to reproduce the
reported outputs.

The manuscript package also supplies figures, numerical summaries, and
plotting scripts for checking reported arithmetic. A complete independent
rerun still requires the task-level run manifests, experiment runner,
benchmark environments, evaluator traces, and trained checkpoint files;
offline selector-training cost is not included in the reported task-run
token totals.

\bibliography{iclr2027_conference}
\bibliographystyle{iclr2027_conference}

\clearpage
\appendix
\clearpage
\begingroup
\hypersetup{linkcolor=black}
\color{black}
\small
\setlength{\parindent}{0pt}
\setlength{\parskip}{0.15ex}
\newcommand{\AppContentsSection}[2]{%
  \noindent\hyperref[#1]{\textbf{\ref*{#1}\quad #2}}\dotfill
  \hyperref[#1]{\pageref*{#1}}\par}
\newcommand{\AppContentsSubsection}[2]{%
  \noindent\hspace*{1.5em}\hyperref[#1]{\ref*{#1}\quad #2}\dotfill
  \hyperref[#1]{\pageref*{#1}}\par}

\begin{center}
{\large\bfseries APPENDIX CONTENTS}
\end{center}
\vspace{0.5em}

\AppContentsSection{app:limitations}{Limitations and Future Work}

\AppContentsSection{app:math}{Memory Construction and Retrieval Rules}
\AppContentsSubsection{app:admission}{Verification and Admission}
\AppContentsSubsection{app:record_example}{What a Record Contains}
\AppContentsSubsection{app:capacity_reuse}{Capacity and Reuse}
\AppContentsSubsection{app:retrieval_selection}{From Stored Records to a Selection}

\AppContentsSection{app:selector}{Reinforcement Learning for Memory Use}
\AppContentsSubsection{app:rl_observation}{Decision Process and Observations}
\AppContentsSubsection{app:reward}{Reward and Q-Learning}
\AppContentsSubsection{app:algorithm}{Construction, Training, and Evaluation Procedure}

\AppContentsSection{app:protocol}{Experimental Details}
\AppContentsSubsection{app:benchmarks}{Benchmarks and Evaluators}
\AppContentsSubsection{app:heldout}{Held-Out SQL Configuration}
\AppContentsSubsection{app:accumulation_protocol}{Accumulation Checkpoints and Record Retention}
\AppContentsSubsection{app:transfer_protocol}{Transfer Conditions and Reuse Measurements}
\AppContentsSubsection{app:baselines}{Baselines and Interventions}
\AppContentsSubsection{app:metric_conventions}{Metric Conventions}
\AppContentsSubsection{app:taskvalues}{Values Underlying the Benchmark Breakdown}
\AppContentsSubsection{app:prompts}{Prompt and Parser Interfaces}

\AppContentsSection{app:additional}{Additional Analyses}
\AppContentsSubsection{app:backbone}{Backbone Summary}
\AppContentsSubsection{app:fixedbudget}{Comparison at a Common 6M-Token Budget}
\AppContentsSubsection{app:rescue}{Rescued Tasks, Broken Tasks, and Applicability Conditions}
\endgroup
\clearpage

\section{Limitations and Future Work}
\label{app:limitations}

\textbf{What verification establishes.}
The method requires executable checks and a way to reset or copy the state
before testing a local edit. A test suite may omit cases, and a reference
check may cover only one input. A checked improvement supports using the correction in that setting;
it does not prove correctness on every future task. Small edits can also miss failures that require a different plan.

\textbf{What memory can represent and reuse.}
A record can become stale when tools, schemas, or requirements change.
The transfer experiments show that a correction valid at its source can
harm a different task. The selector sees compressed features and chooses
at most one record, so it cannot directly combine several records into a
single correction.

\textbf{What the experiments establish.}
The comparisons evaluate the complete method. They do not isolate DQN's
benefit against a fixed selector. The appendix specifies the principal
reported-run configuration, while exact task manifests, evaluator traces,
and execution environments remain separate run artifacts
(\LinkAppendix{app:reproducibility}). Total task-run tokens, calls per solved
task, and retrieval overhead measure different costs; none alone
establishes lower total training cost, runtime, financial cost, or safe
deployment.

\textbf{Future work.}
The limitations above motivate three direct extensions: combining multiple
verified records when one correction is insufficient, detecting or retiring
stale records when tools or task requirements change, and reporting
end-to-end training and runtime costs alongside task-run interaction metrics.

\section{Memory Construction and Retrieval Rules}
\label{app:math}

\subsection{Verification and Admission}
\label{app:admission}
This appendix expands the admission, retention, and retrieval steps.
Admission decides whether to save a newly checked comparison; retention
decides what to remove when space is needed; retrieval finds candidates
for a current decision. The selector then decides whether to use one.

For the original action's outcome $o=H(s,a)$ and an edited action's outcome $o'=H(s,a')$,
let $\Delta=U(o')-U(o)$. Both actions are evaluated from the same state.
A record can be admitted only if the candidate evaluation completes and
$\Delta>\epsilon_{\mathrm{adm}}\geq0$. The reported runs use $\epsilon_{\mathrm{adm}}=0.05$, applied to
the raw utility gain $\Delta$ before any normalization. With binary
correctness this means replacing a failed outcome with a successful one;
partial-credit improvements need not solve the full task. The record is
\begin{equation}
 m_i=(q_i,a_i^{-},a_i^{+},e_i,c_i,u_i).
 \label{eq:app_record}
\end{equation}
Its source-task identifier is retained as metadata. The first five fields
describe the situation, original action, improved action, check results, and
conditions for using the correction. The statistic $u_i$ starts at zero
and summarizes reuse outcomes observed while construction remains active.
It is frozen with the record before selector training and held-out testing.

Among records that pass verification, this score determines which to save:
\begin{equation}
 S_{\mathrm{adm}}(m_i)=
 \lambda_{\Delta}\widetilde{\Delta}_i+
 \lambda_u\widetilde{u}_i-
 \lambda_r R(m_i,\mathcal M)-\lambda_I I(m_i,\mathcal M),
 \label{eq:app_admission}
\end{equation}
where tildes denote normalized quantities, $R$ measures repeated advice,
and $I$ measures conflicting advice. The score favors larger checked gains
and helpful reuse, and penalizes repetition and conflict. A record enters
only if its score exceeds $\tau_{\mathrm{adm}}$. An unverified record
cannot enter, however useful it might appear.

\subsection{What a Record Contains}
\label{app:record_example}
\LinkTable{tab:record_example} expands the running SQL example. It is an
illustration of the record format, not an additional measured test case.
A condition states when to try the correction. Matching that wording does
not guarantee that the correction will solve a future task.

\begin{table}[ht]
\centering
\caption{Illustrative memory record for an inclusive age boundary.}
\label{tab:record_example}
\small
\begin{tabularx}{\linewidth}{lX}
\toprule
Field & Content and purpose\\
\midrule
$q$: situation & A numeric threshold predicate may omit its boundary value.\\
$a^{-}$: inferior action & The original query uses \texttt{age > 18}.\\
$a^{+}$: improved action & Replace that predicate with \texttt{age >= 18}.\\
$e$: evidence & The original and edited queries' result-check outcomes, from the same database state.\\
$c$: conditions & The question requests ``at least'' the threshold; the column and comparison types are compatible.\\
$u$: reuse statistic & Initially zero; later summarizes observed helpful, neutral, or harmful reuse during construction.\\
Source identifier & Identifies the task that supplied the comparison, allowing same-task retrieval to be excluded.\\
\bottomrule
\end{tabularx}
\end{table}

\subsection{Capacity and Reuse}
\label{app:capacity_reuse}
Capacity $B$ is a limit on the number of correction records, not on
source tasks, tokens, or model parameters. The store holds at most $B$
records; processing another task need not add a record if no candidate
passes verification and admission. When capacity is exceeded, retention
ranks records by
\begin{equation}
 S_{\mathrm{keep}}(m_i)=\alpha\widetilde{\Delta}_i+
 \beta\widetilde{u}_i+\omega F_i-\eta R(m_i,\mathcal M),
 \label{eq:app_keep}
\end{equation}
Here $\widetilde{\Delta}_i$ measures improvement on the source action,
$\widetilde{u}_i$ summarizes later reuse outcomes, $F_i$ is normalized reuse
frequency, and $R$ penalizes redundant records. The coefficients $\alpha$,
$\beta$, $\omega$, and $\eta$ weight improvement, reuse utility, frequency,
and redundancy. The reported utility-retention runs use
$(\alpha,\beta,\omega,\eta)=(1.0,1.0,0.5,0.7)$. The lowest-ranked records
are removed.
During construction, when records can still be updated, the score
summarizing a record's reuse is updated as
\begin{equation}
 u_i^{(j+1)}=(1-\rho)u_i^{(j)}+\rho y_{i,j},
 \label{eq:app_reuse_update}
\end{equation}
Here $j$ indexes reuse observations and $y_{i,j}\in\{-1,0,+1\}$ labels
harmful, neutral, or beneficial reuse. In the reference implementation,
this label is assigned relative to the memory-free draft at the same
decision. If the executed action is unchanged from that draft, the reuse is
neutral. If memory changes the action and the task is solved, the reuse is
beneficial. Otherwise, when a previous evaluated attempt is available, the
label follows the sign of the utility change relative to that attempt; if no
such reference is available, the label is neutral. This definition avoids
penalizing a retrieved record merely because it was shown on a difficult
task without changing the submitted action.
Records and these statistics remain unchanged in held-out evaluation.
The three scores answer different questions. $\Delta_i$ asks whether the
edit improved its original action. $u_i$ summarizes what happened when the
saved correction was reused during construction. The DQN value
$Q_\theta$ predicts the reward from choosing that record at the current
decision; training this prediction leaves $\Delta_i$ and $u_i$ fixed.

\subsection{From Stored Records to a Selection}
\label{app:retrieval_selection}
For task $x$ and state $s$, define eligible records as
\begin{equation}
 \mathcal C(x,s)=\{m_i\in\mathcal M:
 x_i^{\mathrm{source}}\ne x,\ c_i\text{ is compatible with }s\}.
\end{equation}
Retrieval selects up to $k$ eligible records using
\begin{equation}
 r_i(s)=\operatorname{sim}(\phi(s),\phi(q_i))+\lambda_{\mathrm{reuse}}u_i,
 \qquad
 \mathcal R_k=\operatorname{TopK}_{m_i\in\mathcal C(x,s)} r_i(s),
 \label{eq:app_retrieval}
\end{equation}
where $\phi$ represents text for similarity comparison. The first term
matches the current situation to the record; the second favors records
whose past use helped, with weight $\lambda_{\mathrm{reuse}}$.
The ordered list $\mathcal R_k$ is passed to the selector as $\mathcal R_t$.
The selector $\mu_\theta$ gives the LLM one record or none
(\LinkSection{sec:memory_rl}); the evaluator checks the resulting action.

\paragraph{Reference implementation details.}
The generator uses the same backbone as the task-solving agent rather than
a separate model. Counterfactual generation is triggered only after a failed
decision while verification budget remains, with at most two failed decisions
expanded per task. The implementation first asks the model for local edits and
uses rule-based edit operators only to fill unused candidate slots up to
$K=4$. Thus the generator changes one decision unit at a time; it does not
replace the complete solution with an independently generated submission.

For retrieval, the implementation embeds the record situation $q_i$ and the
current task prompt plus state, and ranks them with cosine similarity before
adding the reuse term in \LinkEquation{eq:app_retrieval}. The default embedder is
a 512-dimensional character $n$-gram hashing representation. This is an
implementation choice for the reported system, not a claim that this
representation is uniquely required by the method.

The specified defaults refer to different stages:
\begin{center}
\begin{tabular}{lll}
\toprule
Symbol & Value & What it limits\\
\midrule
$K$ & 4 & Alternatives proposed for one failed action\\
$B$ & 200 & Correction records retained in the store\\
$k$ & 3 & Records returned at one retrieval step\\
$L$ & 4 & Candidate slots in the selector architecture\\
\bottomrule
\end{tabular}
\end{center}
Only one retrieved record can be used at a decision. With $k=3$, the fourth
architectural slot is masked; it does not create a fourth retrieved record.
Admission weights in
\LinkEquation{eq:app_admission} are $(1.0,0.5,0.7,1.0)$, with threshold
$\tau_{\mathrm{adm}}=0.1$; the raw improvement floor is $\epsilon_{\mathrm{adm}}=0.05$;
the reuse update uses $\rho=0.3$; and retrieval uses
$\lambda_{\mathrm{reuse}}=0.2$. The utility-retention coefficients are
$(\alpha,\beta,\omega,\eta)=(1.0,1.0,0.5,0.7)$. Construction checks have
a separate verification allowance from ordinary solving; both enter
interaction accounting. These values are fixed in the reported runs.

The exact normalization functions used inside \LinkEquation{eq:app_admission}
and \LinkEquation{eq:app_keep}, together with the internal redundancy and conflict
computations, are implementation details of the run code rather than new
measured quantities. They do not change the definitions of $\Delta$, $u$,
or the eligibility rule in Eq.~(7).

\section{Reinforcement Learning for Memory Use}
\label{app:selector}

\subsection{Decision Process and Observations}
\label{app:rl_observation}
The selector chooses which record, if any, the LLM sees before it acts.
The LLM still writes the task action. An episode is one task.
At each decision, the frozen LLM drafts an action
$\hat a_t$, retrieval supplies $\mathcal R_t$, and the memory-use policy
chooses $j_t\in\mathcal J_t=\{0,\ldots,|\mathcal R_t|\}$.
Action zero executes the draft. Action $j_t>0$ presents only record
$\mathcal R_t[j_t]$ and asks the LLM to revise the draft. The resulting
action is checked by $H$, producing the next agent state and cost counters.
The policy does not choose an arbitrary subset or modify the verifier.

The selector receives a numerical description $z_t$ of the current
situation, draft action, available records, and remaining budgets:
\begin{equation}
 z_t=\big[P E(s_t)\,\Vert\,P E(\hat a_t)\,\Vert\,
 f_{t,1}\,\Vert\cdots\Vert\,f_{t,L}\,\Vert\,b_t\big].
 \label{eq:rl_observation}
\end{equation}
Here, $\Vert$ joins the feature vectors. $E$ converts text to vectors
(it is denoted $\phi$ in the retrieval rule), and the fixed mapping $P$
reduces each text vector to 32 dimensions. In the reference implementation,
$P$ is a fixed-seed Gaussian random projection from the 512-dimensional
retrieval embedding to 32 dimensions (seed 20260824). It is not fitted during
memory construction or selector training. The vector $b_t$ describes the
remaining budgets, using normalized values from the execution log.
Each seven-dimensional record vector contains
relevance to the state, admission score, estimated improvement from its
evidence, reuse utility, use frequency, helpful-use fraction, and validity.
Features are normalized, and an unused helpful-use fraction is set to zero.
The specified controller has $L=4$ candidate slots and six budget features,
giving $32+32+4\times7+6=98$ dimensions. Retrieval may return fewer
records: the unused slots are zero-padded and masked, including the fourth
slot when $k=3$. Slot capacity $L$, retrieval count $k$, and generation
budget $K$ are separate quantities.

The Q-network is a two-layer multilayer perceptron with hidden widths
$(64,64)$ and ReLU activations. This architecture maps the 98-dimensional
observation to one value for the skip action and one value for each
architectural record slot.

This compressed observation does not expose the complete agent history.
The controller therefore learns under partial observation; we do not assume
that $z_t$ is a fully observed Markov state. Frozen LLM parameters also do
not imply deterministic action generation.

\subsection{Reward and Q-Learning}
\label{app:reward}\label{app:rl_objective}
At decision $t$, the selector receives
\begin{equation}
r_t=\mathbf{1}[t=T]\mathbf{1}[\mathrm{solved}]
-\lambda_w\delta_{w_t}-\lambda_f\delta_{f_t}
-\lambda_\tau\delta_{\tau_t}/\tau_0.
\label{eq:reward}
\end{equation}
$T$ is the final decision in the task. The first term pays one unit only
when the task ends successfully. The nonnegative increments
$\delta_{w_t}$, $\delta_{f_t}$, and $\delta_{\tau_t}$ count evaluator calls,
failed evaluated actions, and tokens charged to that decision. The
$\lambda$ coefficients weight the penalties, and $\tau_0$ sets the token
scale. This reward favors completing the task while spending less on
unsuccessful attempts and unnecessary use of memory.

The cost increments are positive changes in the charged counters; reducing
cost avoids a penalty rather than generating a separate reward. Token normalization
prevents thousands of tokens from overwhelming the call penalty solely
because of their unit. The local gain $\Delta$ determines eligibility for storage;
adding it again as an RL reward would reward source-task improvement
without establishing useful reuse.

The DQN assigns each choice $j$ a score: its expected total future reward,
with rewards farther in the future weighted by powers of $\gamma$:
\begin{equation}
 Q_\theta(z,j)\approx
 \mathbb E\!\left[\sum_{\ell=0}^{T-t}\gamma^\ell r_{t+\ell}
 \mid z_t=z,j_t=j\right].
 \label{eq:q_return}
\end{equation}
Training saves each choice and its outcome as
$(z_t,j_t,r_t,z_{t+1},d_t)$, where $d_t=1$ marks an episode's end.
A periodically updated copy of the Q-network, with parameters $\theta^-$,
provides the learning target:
\begin{align}
 y_t&=r_t+\gamma(1-d_t)
 \max_{j\in\mathcal J_{t+1}}Q_{\theta^-}(z_{t+1},j),
 \label{eq:dqn_target}\\
 \mathcal L(\theta)&=\mathbb E_{\mathcal D}
 \big[(y_t-Q_\theta(z_t,j_t))^2\big].
 \label{eq:dqn_loss}
\end{align}
$\mathcal D$ is the replay buffer of saved training steps. The loss reduces
the difference between the predicted score and target $y_t$.
Terminal transitions have target $r_t$; unavailable record slots cannot
be selected. During training, $\varepsilon$-greedy selection occasionally
tries a random valid choice instead of the highest-scoring one.
Held-out evaluation uses $j_t=\arg\max_{j\in\mathcal J_t}Q_\theta(z_t,j)$.
The replay buffer stores training transitions; it is distinct from the
persistent store of correction records. Only $\theta$ is optimized: the base LLM, generator, evaluator, and fixed
admission rules are not trained by this objective.

\begin{table}[ht]
\centering
\caption{DQN configuration used for the reported memory-use selector.}
\label{tab:rl_configuration}
\begin{tabular}{ll}
\toprule
Setting & Value\\
\midrule
Training decision steps & 2,000\\
Hidden layers & $64,64$\\
Activation & ReLU\\
Optimizer & Adam, $\epsilon_{\mathrm{Adam}}=10^{-8}$\\
Learning rate & $10^{-3}$\\
Learning starts & 100 decisions\\
Training frequency & every 4 decisions\\
Discount $\gamma$ & 0.95\\
Replay capacity & $10^4$ transitions\\
Batch size & 32\\
Target synchronization interval & 250 steps\\
Exploration $\varepsilon$ & 0.30 to 0.05\\
Call penalty $\lambda_w$ & 0.15\\
Failure penalty $\lambda_f$ & 0.25\\
Token penalty $\lambda_\tau$ & 0.02\\
Token scale $\tau_0$ & $10^4$ tokens\\
Selector seeds & $0,1,2,3,4$\\
\bottomrule
\end{tabular}
\end{table}
\LinkTable{tab:rl_configuration} counts decisions, not tasks. For each
selector seed, candidate checkpoints are evaluated every 250 training
decisions on a held-out selector-validation subset drawn from
$\mathcal T_{\mathrm{train}}$. The checkpoint with the highest mean return
on that subset is used for evaluation. The validation subset remains
disjoint from $\mathcal T_{\mathrm{eval}}$.

\subsection{Construction, Training, and Evaluation Procedure}
\label{app:algorithm}
Let $\mathcal T_{\mathrm{build}}$, $\mathcal T_{\mathrm{train}}$, and
$\mathcal T_{\mathrm{eval}}$ denote construction, RL training, and evaluation
tasks. The held-out protocol requires
\begin{equation}
 (\mathcal T_{\mathrm{build}}\cup\mathcal T_{\mathrm{train}})
 \cap\mathcal T_{\mathrm{eval}}=\varnothing.
 \label{eq:rl_split}
\end{equation}
Construction and training may share task identifiers under this condition;
both must exclude all evaluation identifiers. Retrieval separately excludes
a record whenever its source identifier equals the current task identifier,
including during training. Memory construction
collects and admits verified records; training then learns which records
to use while keeping that saved store fixed. If additional source tasks are used to
update the store, the result is a new snapshot even if its record count
stays at the capacity limit. During held-out
evaluation, neither records nor features derived from reuse history change.
Evaluation logs may count uses, but must not feed those counts back into
retrieval or the selector. Policy parameters are frozen as well.

\begin{algorithm}[ht]
\caption{Verified memory construction, RL training, and held-out reuse}
\label{alg:cfmemory}
\begin{algorithmic}[1]
\Require Fixed LLM $\pi$, generator $G$, evaluator $H$; budgets $K,B,k$;
 task sets $\mathcal T_{\mathrm{build}},\mathcal T_{\mathrm{train}},\mathcal T_{\mathrm{eval}}$
\State Initialize store $\mathcal M$, Q-network $Q_\theta$, target $Q_{\theta^-}$, replay $\mathcal D$
\State \textbf{Construct memory on $\mathcal T_{\mathrm{build}}$:}
\For{each observed failed decision $(x,s,a,o)$ with verification budget}
 \State Generate up to $K$ local alternatives $G(s,a,o;K)$
 \For{each alternative $a'$ while verification budget remains}
  \State Evaluate $o'\gets H(s,a')$ from a copy or reset of $s$
  \If{evaluation completes and $U(o')-U(o)>\epsilon_{\mathrm{adm}}$}
   \State Distill $(q,a^-,a^+,e,c,u)$ with source identifier $x$
   \State Apply admission and retention rules; keep $|\mathcal M|\le B$
  \EndIf
 \EndFor
\EndFor
\State Freeze memory snapshot and reuse features
\State \textbf{Train selection on $\mathcal T_{\mathrm{train}}$:}
\For{each task until the training-step budget is exhausted}
 \For{each decision until the task terminates or its budget is exhausted}
  \State Draft $\hat a\sim\pi(\cdot\mid s)$; retrieve $\mathcal R$ excluding this task's records
  \State Encode $z$; choose valid $j$ by $\varepsilon$-greedy $Q_\theta$
  \State Use draft if $j=0$; otherwise revise it with record $\mathcal R[j]$
  \State Execute action with $H$; obtain reward $r$, next observation $z'$, and terminal flag $d$
  \State Add $(z,j,r,z',d)$ to $\mathcal D$; minimize \LinkEquation{eq:dqn_loss} on a batch
  \State Synchronize $\theta^-\gets\theta$ at the target-update interval
 \EndFor
\EndFor
\State \textbf{Evaluate on $\mathcal T_{\mathrm{eval}}$:}
\State Freeze $\mathcal M$ and $\theta$; repeat the decision loop with greedy selection and no updates
\end{algorithmic}
\end{algorithm}

\section{Experimental Details}
\label{app:protocol}

\subsection{Benchmarks and Evaluators}
\label{app:benchmarks}
A benchmark supplies tasks; an evaluator checks attempted solutions. The
decision unit is the part of a solution that the agent changes, not
necessarily a complete task. The twelve settings below provide executable
checks for candidate actions.
SQL execution is combined with the benchmark's result checker; successful
execution alone does not establish that a query answers the question.

\begin{table*}[t]
\centering
\caption{Evaluation suite and executable checker used as the world model.}
\footnotesize
\setlength{\tabcolsep}{4.0pt}
\resizebox{\textwidth}{!}{%
\begin{tabular}{lllcl}
\toprule
\textbf{Domain} & \textbf{Benchmark} & \textbf{Evaluator} & \textbf{Score} & \textbf{Decision unit} \\
\midrule
Math & miniF2F \citep{zheng2022minif2f} & Lean kernel & pass rate & tactic / proof step \\
Math & ProofNet \citep{azerbayev2023proofnet} & Lean kernel & pass rate & proof continuation \\
Math & AMBER \citep{yang2026amber} & Lean-based checker & pass rate & tactic / lemma choice \\
Coding & SWE-bench Verified \citep{jimenez2024swebench,openai2024swebenchverified} & repository tests & resolved rate & edit / patch \\
Coding & BugsInPy \citep{widyasari2020bugsinpy} & regression tests & repair rate & localized edit \\
Formal & DafnyBench \citep{loughridge2024dafnybench} & Dafny verifier & verification & invariant / annotation \\
Formal & Vericoding-Dafny \citep{bursuc2025vericoding} & Dafny verifier & verified synthesis & code step \\
Formal & Vericoding-Verus \citep{bursuc2025vericoding} & Verus verifier & verified synthesis & proof / code step \\
Text-to-SQL & Spider \citep{yu2018spider} & SQL engine & exec. accuracy & join / predicate \\
Text-to-SQL & BIRD \citep{li2023bird} & SQL engine & exec. accuracy & schema / join \\
SMT & SMT-LIB / SMT-COMP \citep{barrett2016smtlib,smtcomp} & SMT solver & solved rate & encoding / solver action \\
SAT & SAT Competition / SATLIB \citep{satcompetition,hoos2000satlib} & SAT solver & solved rate & branching / encoding \\
\bottomrule
\end{tabular}%
}
\end{table*}

\subsection{Held-Out SQL Configuration}
\label{app:heldout}
The held-out SQL configuration uses 60 construction tasks and 40 evaluation
tasks, sampled with seeds $12345+8675309$ and 12345, respectively. Task
identifiers shared by the two sets are removed, and the sets are checked
for overlap before memory is frozen.
The configuration that also separates databases contains 35 construction databases and 34
evaluation databases with no overlap. These are configuration details, not
additional outcome measurements in the main table.

The agent observes the question, database identifier, and optional evidence.
The verifier returns reference-derived correctness feedback without revealing
the reference SQL. Evaluation disables record construction, admission,
eviction, and reuse-score updates; the low-utility suppression gate is also
inactive. The memory-use policy is trained on a separate training stream
and is fixed during evaluation (\LinkAppendix{app:selector}).
The construction/evaluation counts above do not specify the size of that
RL training stream; 2,000 decision steps should not be read as 2,000 tasks.

\subsection{Accumulation Checkpoints and Record Retention}
\label{app:accumulation_protocol}

\textbf{What accumulates.}
As the agent attempts source tasks, it tests edits to failed actions and
saves corrections that pass verification and admission. The store accumulates
\emph{accepted correction records}, not raw tasks or new LLM parameters.
A source task can add no record or several. A checkpoint is a saved store
after a specified number of source tasks have been processed. The plotted
checkpoints are 0, 25, 50, 100, and 200 tasks. At zero source tasks, the
store contains no corrections from that construction stream.

\textbf{How to read the panels.}
Panel (a) compares held-out performance across construction checkpoints
within each method and between \method and ReasoningBank. Evaluation
reads that snapshot without adding records or changing reuse statistics.
Panel (b) reports retrieval and context tokens per task, rather than all
construction and solving tokens. Panel (c), labeled regression on earlier
tasks, describes loss of earlier-task performance as construction advances;
it is a different measurement from the held-out score in (a).
The figure marks one standard deviation across five source-task
orderings. The same trained selector is reused unchanged at the
0/25/50/100/200 construction checkpoints; only the memory snapshot changes.
This keeps the accumulation curve focused on additional stored experience
rather than retraining a different selector at each checkpoint.

\textbf{What fixed capacity controls.}
Panel (d) gives each retention rule the same maximum record count. Recency
prioritizes recent records; random retention chooses without a utility or
similarity ranking; similarity uses a similarity ranking; and utility uses
the original checked improvement and outcomes of later use during construction, as in \LinkEquation{eq:app_keep}.
Keeping the record limit equal prevents a method from gaining simply by
being allowed to retain more records. It does not equate record lengths,
retrieval tokens, or all computational costs.

\textbf{Run interpretation.}
The mechanism uses $B=200$ as the record-capacity limit. Checkpoints still
count processed source tasks rather than stored records, so a 200-task
checkpoint need not contain 200 memories. The five task orderings determine
the mean and standard deviation shown in the figure, while the fixed
selector policy described above is shared across checkpoints. Panel (c)
remains a regression measurement on the designated earlier-task set rather
than another held-out-score curve. The figure is therefore not evidence of
unbounded growth in the number of memories.

\subsection{Transfer Conditions and Reuse Measurements}
\label{app:transfer_protocol}

Transfer asks whether a correction learned on a source task remains useful
when some part of the next task changes. An environment is the setting in
which actions are executed, such as a database and schema, repository, or
verifier configuration. The examples in \LinkTable{tab:transfer_design}
illustrate the intended distinctions. The reported transfer study uses
60 source--target task pairs for each of the four conditions.

\begin{table}[ht]
\centering
\caption{Meaning of the four transfer conditions, with illustrative examples.}
\label{tab:transfer_design}
\small
\begin{tabularx}{\linewidth}{p{0.23\linewidth}XX}
\toprule
Condition & What changes & Example of the reuse question\\
\midrule
Same environment, new task & A different request within the source environment. & Does the boundary correction help a new query over the same schema?\\
New environment, new task & The execution setting as well as the request. & Can a predicate correction apply in a different database with compatible types?\\
Same error, new surface & The form of the task or action, while the underlying mistake is similar. & Does the record match a renamed column or a differently worded inclusive request?\\
Look-alike, changed condition & Surface similarity remains, but a requirement for the correction fails. & Can the agent avoid changing $>$ to $\geq$ when the request requires a strict boundary?\\
\bottomrule
\end{tabularx}
\end{table}

\textbf{Three separate questions.}
Task success asks whether the complete task passes its evaluator.
Retrieval coverage asks whether retrieval supplies guidance. Harmful use
asks whether acting on memory worsens the outcome; retrieval alone is not
harmful use, and a failed task alone does not establish that memory caused
its failure. A candidate can be retrieved and then skipped by the selector.
The transfer figure's reported coverage should therefore not be interpreted
as either a selection rate or a correctness rate.

\textbf{Denominators and harm labels.}
Retrieval coverage is measured per \emph{retrieval opportunity}: the
numerator is the number of retrieval opportunities for which at least one
eligible record is returned, and the denominator is the number of
opportunities at which retrieval is attempted. Records rejected by the
compatibility filter do not count as coverage.

Harmful use is measured per \emph{selected record use}. The denominator
counts decisions at which the selector chooses a memory record rather than
skipping. The numerator counts those selected uses labeled harmful by the
same decision-level reuse criterion used during construction
(\LinkAppendix{app:math}). Thus a record that is retrieved but skipped is
not a harmful use, and a task failure without a selected harmful action is
not counted as one. The condition-removal percentages use the same
selected-use denominator.

The retrieval intervention in \LinkFigure{fig:components}(d) uses a
different normalization: its action-change and harmful-change rates are
reported per retrieved memory, as stated in that figure. The rescued/broken
experiment in \LinkAppendix{app:rescue} is different again, using 120
paired tasks per domain. These three denominators should not be
interchanged.

\subsection{Baselines and Interventions}
\label{app:baselines}
In each main-table group, the agent procedure and base LLM are fixed;
the rows change the added mechanism. ExpeL extracts lessons from past attempts; Generative Agents combines saved
experiences and reflections; Voyager retains reusable procedures; MemGPT
manages information between context and external storage. These components
are adapted to the common task interface. Best-of-$N$
generates additional current-task candidates without persistent memory.
ReasoningBank extracts strategies from past attempts. In the separate
6M-token comparison, MaTTS gives it more trial experience from which to
learn; the verified-feedback variant gives its memory-building process
checked outcomes. These controls ask whether either addition brings
ReasoningBank to the full method's score. They compare complete methods
and do not isolate each difference in how records are formed or used.

In the component ablation, \emph{CF without WM} proposes memory advice without
testing whether the suggested change works; ordinary task actions are still evaluated.
\emph{WM without CF} tests alternatives for the current task but saves no
corrections for later tasks. In the retrieval ablation, \emph{random}
retrieval samples records without state matching; \emph{shuffled} retrieval
reassigns records to other decisions; \emph{correct} (matched) retrieval uses
the recorded situations and conditions. No retrieval means the LLM receives no memory record. These interventions alter the
records available for selection; they are not a comparison between a
trained selector and a fixed selection rule. The manuscript package alone does not contain complete runnable baseline
configurations. The reference implementation resolves two details for the
Best-of-$N$ compute control: it uses $N=1+K=5$ candidates, and an LLM judge
selects one candidate for execution. The verifier is not used to choose among
those candidates, so this control increases current-task inference rather
than persistent memory or verifier-guided selection. Complete baseline prompts
and per-method construction budgets are still not supplied with the manuscript
package.

\subsection{Metric Conventions}
\label{app:metric_conventions}
A world-model call is one executable evaluation: a Lean check cycle,
repository test invocation, Dafny/Verus attempt, SQL execution with checking,
or solver invocation. A failed attempt reaches that evaluator and fails the
task's success criterion. LLM-only proposals are excluded from these two
counts but included in token accounting. Counterfactual checks are included.

For a benchmark with $n_{\mathrm{solved}}$ solved instances, normalized
interaction counts are $N_{\mathrm{WM}}/n_{\mathrm{solved}}$ and
$N_{\mathrm{fail}}/n_{\mathrm{solved}}$. Reported reductions subtract the
augmented agent's normalized count from its base agent's count. These
ratios can change through both their numerator and denominator; a reduction
does not establish a lower total call count on its own.
The main table reports total task-run tokens, without dividing by solved
tasks. Offline selector-training tokens and training compute are not
separately itemized. A fixed task-run token budget therefore does not
establish equal total training cost or equal wall-clock time.

\subsection{Values Underlying the Benchmark Breakdown}
\label{app:taskvalues}
\begin{figure}[ht]
\centering
\includegraphics[width=\linewidth]{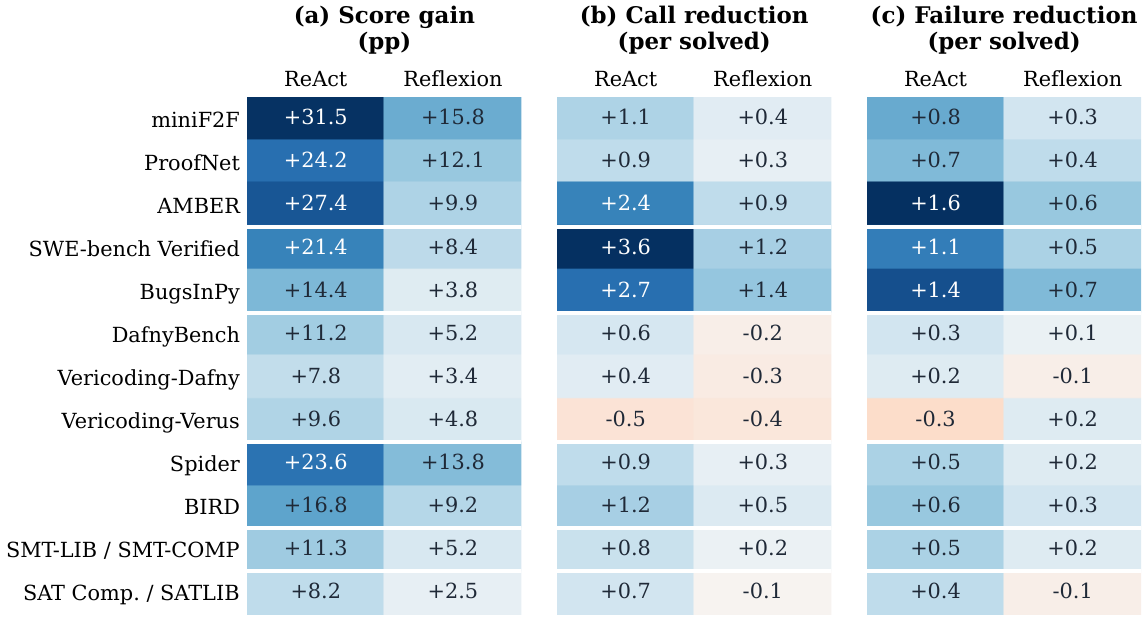}
\caption{\textbf{Adding CF Memory to each base agent on 12 benchmarks.}
The LLM is gpt-oss-120b. Each cell compares the augmented agent with its
own unaugmented version. Positive entries mean higher scores in (a) or
fewer evaluator calls and failed attempts per solved task in (b,c).
Negative reductions mean additional interaction.}
\label{fig:domain}
\end{figure}
\begin{table}[ht]
\centering
\caption{Values in \LinkFigure{fig:domain}. Each entry is ReAct / Reflexion.
Score changes are pp; call and failure reductions are per solved instance.}
\label{tab:domainapp}
\small
\begin{tabular}{lrrr}
\toprule
Benchmark & Score gain & Call reduction & Failure reduction\\
\midrule
miniF2F & 31.5 / 15.8 & 1.1 / 0.4 & 0.8 / 0.3\\
ProofNet & 24.2 / 12.1 & 0.9 / 0.3 & 0.7 / 0.4\\
AMBER & 27.4 / 9.9 & 2.4 / 0.9 & 1.6 / 0.6\\
SWE-bench Verified & 21.4 / 8.4 & 3.6 / 1.2 & 1.1 / 0.5\\
BugsInPy & 14.4 / 3.8 & 2.7 / 1.4 & 1.4 / 0.7\\
DafnyBench & 11.2 / 5.2 & 0.6 / $-0.2$ & 0.3 / 0.1\\
Vericoding-Dafny & 7.8 / 3.4 & 0.4 / $-0.3$ & 0.2 / $-0.1$\\
Vericoding-Verus & 9.6 / 4.8 & $-0.5$ / $-0.4$ & $-0.3$ / 0.2\\
Spider & 23.6 / 13.8 & 0.9 / 0.3 & 0.5 / 0.2\\
BIRD & 16.8 / 9.2 & 1.2 / 0.5 & 0.6 / 0.3\\
SMT-LIB / SMT-COMP & 11.3 / 5.2 & 0.8 / 0.2 & 0.5 / 0.2\\
SAT Competition / SATLIB & 8.2 / 2.5 & 0.7 / $-0.1$ & 0.4 / $-0.1$\\
\bottomrule
\end{tabular}
\end{table}

The mean over all 24 score changes is 12.5625 pp. Averaging only the four
main-table domains, with equal domain weight, gives 18.5 pp for ReAct and
8.175 pp for Reflexion. This domain-weighted summary differs from the 12-benchmark mean because
domains contain different numbers of benchmarks.

\subsection{Prompt and Parser Interfaces}
\label{app:prompts}

The reported runs use fixed prompts with parser-checked output contracts.
The contracts below state the behavior that affects the experiments; they
do not give the LLM permission to bypass the evaluator or alter more of the
action than the corresponding module allows.

\begin{table*}[t]
\centering
\caption{Prompt interfaces used by the reported runs.}
\label{tab:prompt_interfaces}
\scriptsize
\setlength{\tabcolsep}{3pt}
\renewcommand{\arraystretch}{1.08}
\begin{tabularx}{\textwidth}{>{\raggedright\arraybackslash}p{0.29\textwidth}>{\raggedright\arraybackslash}p{0.18\textwidth}>{\raggedright\arraybackslash}X}
\toprule
Template & Used in & Required behavior\\
\midrule
\makecell[l]{\texttt{GENERATE\_LOCAL}\\\texttt{ALTERNATIVES}} & Module I &
Return local edits in the fixed edit grammar, one changed decision unit per
candidate; do not generate a complete replacement submission.\\
\texttt{DISTIL\_RECORD} & Module II &
Distill the checked contrast into a reusable record. The condition must be
checkable on a future task and must be able to be false on a look-alike task.\\
\texttt{DRAFT\_ACTION} & Module III &
Produce the action draft before any memory is shown. This is the
memory-free draft used by the selector and reuse-credit logic.\\
\texttt{REVISE\_WITH\_RECORD} & Module III &
Show exactly one selected record and revise the draft only when useful; the
model may return the draft unchanged.\\
\makecell[l]{\texttt{RECORD\_RENDER} /\\\texttt{RECORD\_RENDER\_NO}\\\texttt{CONDITION}} &
Condition ablation &
Render the same record fields in both cases; the only intended difference
is whether the applicability clause is shown.\\
\texttt{BEST\_OF\_N\_JUDGE} & Best-of-$N$ control &
Choose exactly one candidate and return \texttt{CHOICE: <n>}. Parse failures
are logged and fall back to candidate 1; the verifier does not choose the
candidate.\\
\texttt{EXPEL\_INSIGHT} & ExpeL baseline &
Extract one transferable lesson from the completed attempt.\\
\makecell[l]{\texttt{GENERATIVE\_AGENTS}\\\texttt{REFLECT}} & Generative Agents baseline &
Write one generalized reflection for later retrieval.\\
\texttt{VOYAGER\_SKILL} & Voyager baseline &
Store a reusable procedure with a name, applicability description, and
steps.\\
\makecell[l]{\texttt{REASONINGBANK}\\\texttt{EXTRACT}} & ReasoningBank baseline &
Produce a strategy with title, description, and content, without adding
COUNTERMEM's condition field.\\
\makecell[l]{\texttt{MATTS\_SELF}\\\texttt{CONTRAST}} & MaTTS control &
Contrast the available attempts to extract strategy-level guidance, without
COUNTERMEM's pairwise executable verification gate.\\
\makecell[l]{\texttt{REASONINGBANK}\\\texttt{VERIFIED}} & Verified-feedback control &
Provide checked step outcomes to the strategy extractor while retaining the
ReasoningBank record format.\\
\texttt{LOOK\_ALIKE\_TASK} & Transfer study &
Generate a superficially similar target for which the stored applicability
condition is false; the generated target is retained only after the
applicability check confirms that the recorded condition does not hold.\\
\bottomrule
\end{tabularx}
\end{table*}

For \texttt{DISTIL\_RECORD}, the condition instruction is:
\begin{quote}\small
\texttt{CONDITION: <a requirement of the NEW task that must hold before applying this.
It must be checkable and it must be able to be FALSE for a task that looks similar.
Write "none" only if no such requirement exists.>}
\end{quote}
The parser maps \texttt{none} to an empty condition, which the compatibility
rule treats as universally applicable. This makes the absence of a
condition explicit rather than storing the literal word as a constraint.

Prompt fields are truncated before rendering to keep token accounting
bounded: state 2500 characters; action or draft 2000; evaluator feedback
800; decision units 4000; environment context 3000; rendered record 1200;
and trajectory 3000.

\section{Additional Analyses}
\label{app:additional}

\subsection{Backbone Summary}
\label{app:backbone}
\begin{figure}[ht]
\centering
\includegraphics[width=\linewidth]{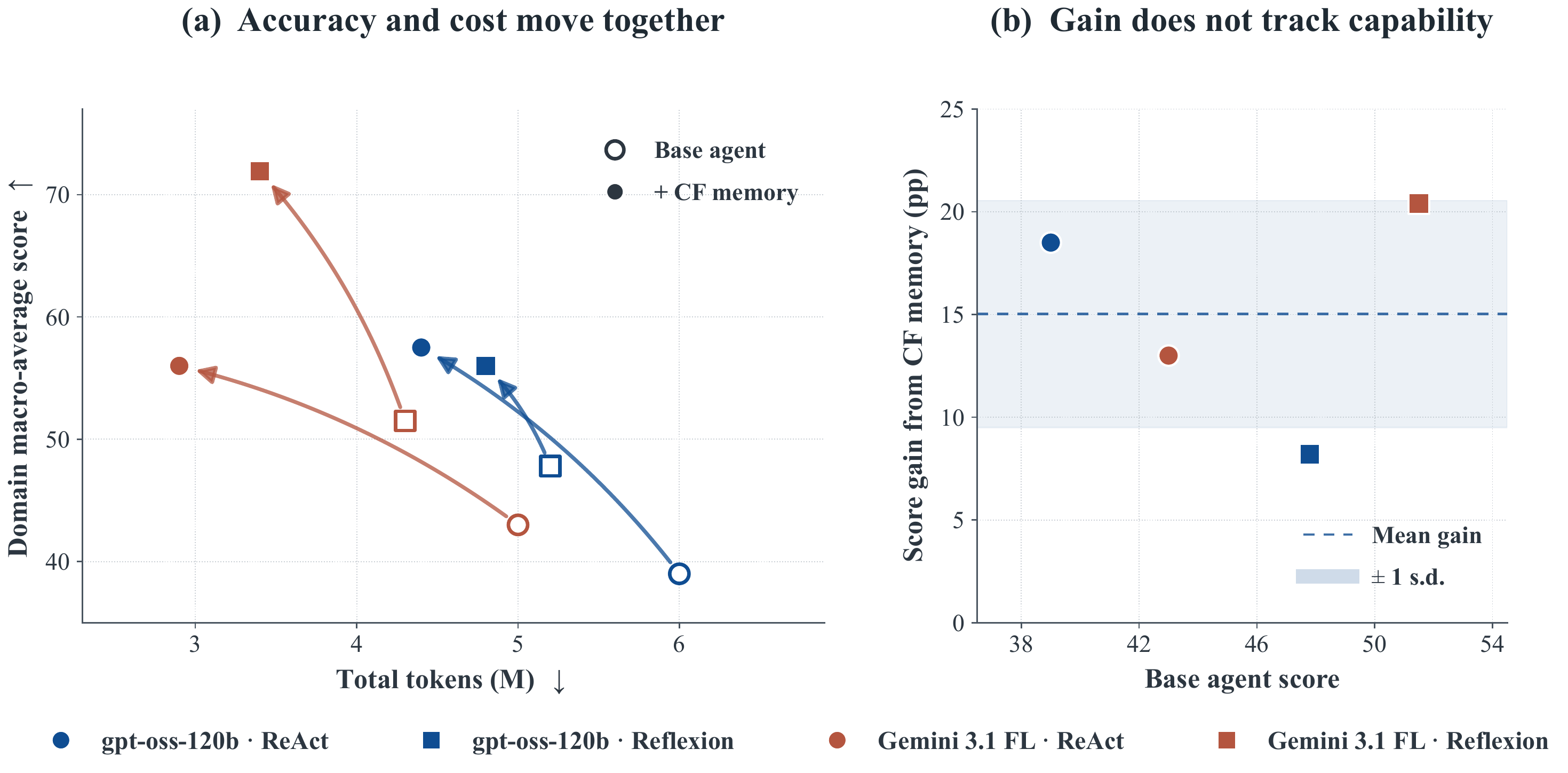}
\caption{\textbf{Four-domain summaries of the main table.}
Each arrow links one unaugmented agent to its CF Memory version. The right
panel plots gain against base score; the shaded band is variation across the
four agent--backbone combinations, not a confidence interval over runs.}
\label{fig:backbone}
\end{figure}
This figure re-expresses the None and CF Memory rows of
\LinkTable{tab:multiagent_multimodel}. All four arrows move toward higher
score and fewer total tokens. Gains are not ordered by the unaugmented
score: gpt-oss-120b ReAct gains 18.5 pp from a score of 39.0, while Gemini
Reflexion gains 20.4 pp from 51.5. Four configurations are insufficient to
establish how memory gains generally depend on the base model's ability.

\subsection{Comparison at a Common 6M-Token Budget}
\label{app:fixedbudget}

To ask whether the score advantage persists when the reported token
allowance is equal, this separate five-domain study compares ReasoningBank,
ReasoningBank with MaTTS \citep{ouyang2026reasoningbank}, our verified-feedback
variant of ReasoningBank, and \method. The MaTTS control asks whether more
trial experience closes the gap; the verified-feedback control asks
whether giving memory construction checked outcomes closes it.
All methods use a common 6M-token task-run
budget. The five domains are Math, Coding, Text-to-SQL, Formal Verification, and
SAT. The budget matches task-run tokens, not offline training cost or
wall-clock time. The package does not specify the run's agent--LLM pair
or how the 6M tokens are allocated among construction and solving.

\begin{figure}[ht]
\centering
\includegraphics[width=\linewidth]{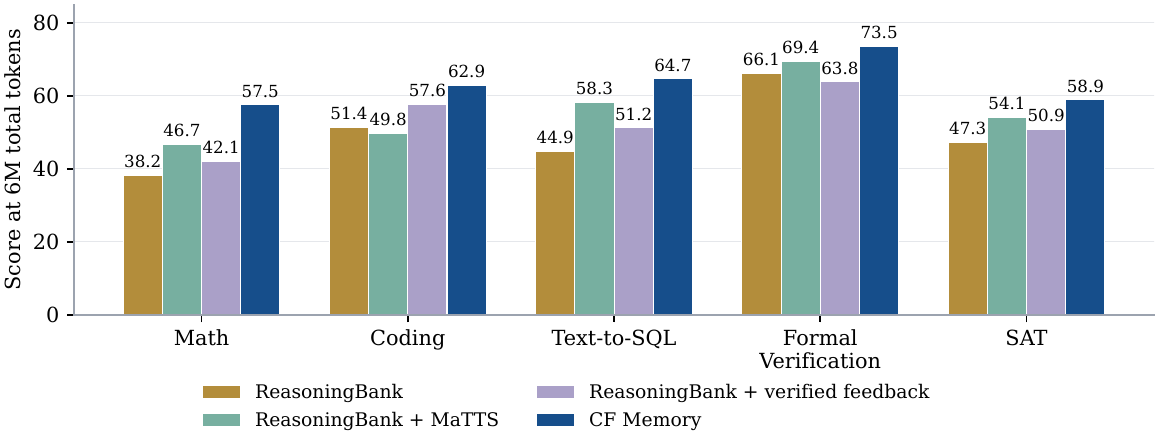}
\caption{\textbf{Comparison at a common 6M-token budget.}
CF Memory outperforms ReasoningBank, ReasoningBank with MaTTS, and
ReasoningBank with verified feedback in all five domains.
Reported averages give each domain equal weight.}
\label{fig:reasoningbank}
\end{figure}

\textbf{Neither control closes the gap.}
\method averages 63.5, exceeding ReasoningBank
with MaTTS by 7.8 pp and ReasoningBank with verified feedback by 10.4 pp
(\LinkFigure{fig:reasoningbank}).

On Formal Verification, giving ReasoningBank checked outcomes changes its
score from 66.1 to 63.8; \method reaches 73.5. Checked outcomes alone therefore
do not explain the gap. The methods also store and use memory differently,
so this comparison neither identifies the decisive difference nor establishes
that verified feedback generally hurts performance.

\subsection{Rescued Tasks, Broken Tasks, and Applicability Conditions}
\label{app:rescue}
For each method and domain, the experiment pairs runs on 120 tasks with
and without that method's memory. A \emph{rescued} task fails without
memory and succeeds with it. A \emph{broken} task succeeds without memory
and fails with it. Tasks whose success status is unchanged contribute to
neither count. The difference, rescued minus broken, measures the net
change in successful tasks within these paired runs.

\begin{figure}[ht]
\centering
\includegraphics[width=\linewidth]{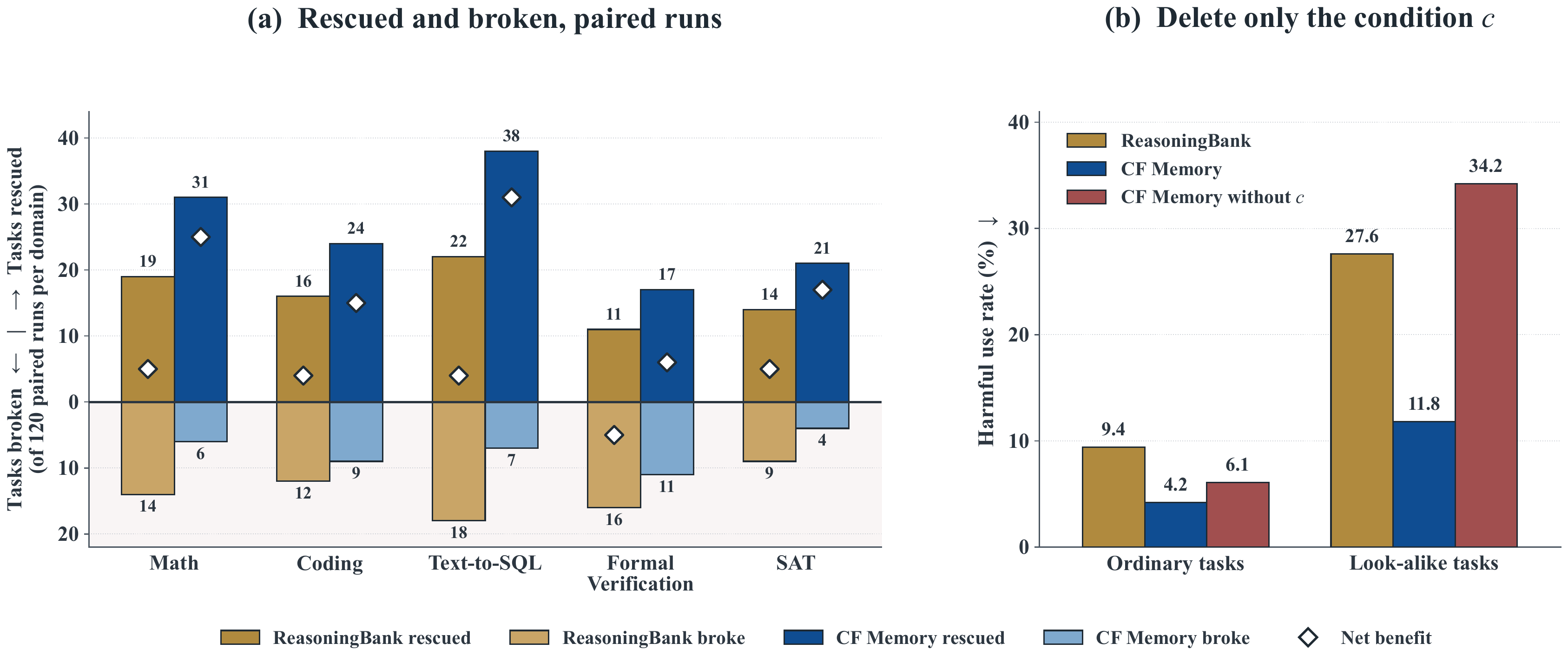}
\caption{\textbf{Benefits and harm from reuse.}
(a) In 120 paired runs per domain, rescued tasks fail without memory and
succeed with it; broken tasks have the opposite outcome. Diamonds mark the
net count, rescued minus broken. (b) Removing only the applicability field
$c$ increases harmful use, especially on look-alike tasks.}
\label{fig:rescue}
\end{figure}
Comparing each method with its own no-memory runs, the net counts for
ReasoningBank are $+5,+4,+4,-5,+5$ across Math, Coding,
SQL, Formal Verification, and SAT. CF Memory yields $+25,+15,+31,+6,+17$.
Across 600 paired runs per method, these sum to 13 and 94 net additional
successes, respectively. CF Memory still breaks tasks: its Formal
Verification results include 17 rescued and 11 broken cases. The aggregate
gain therefore does not mean that memory helps every task.

A second intervention removes only the applicability field $c$ from
the CF mechanism. In the reported ablation, removal is applied at both
sites where $c$ is used: the retrieval filter no longer enforces the
condition, and the rendered record no longer shows the applicability
clause. All other record fields and selector settings are unchanged.
Removing $c$ raises harmful use from 4.2\% to 6.1\% on ordinary tasks and
from 11.8\% to 34.2\% on look-alike tasks. The larger increase on
look-alike tasks supports keeping $c$: stating and enforcing when a
correction applies helps avoid repeating it under the wrong conditions.
These percentages are normalized per selected record use
(\LinkAppendix{app:transfer_protocol}); they are not the per-domain
broken-task rates from the paired experiment.

\end{document}